\documentclass[conference]{IEEEtran}
\usepackage{cite}
\usepackage{amsmath,amssymb,amsfonts}
\usepackage{algorithmic}
\usepackage{graphicx}
\usepackage{textcomp}
\usepackage{xcolor}
\usepackage{mathptmx}
\def\BibTeX{{\rm B\kern-.05em{\sc i\kern-.025em b}\kern-.08em
    T\kern-.1667em\lower.7ex\hbox{E}\kern-.125emX}}
\begin{document}

\title{AIS Data-Driven Maritime Monitoring Based on Transformer: A Comprehensive Review 
}


\author{\IEEEauthorblockN{Zhiye Xie\textsuperscript{1,2}, Enmei Tu\textsuperscript{2,†}, Xianping Fu\textsuperscript{1}, Guoliang Yuan\textsuperscript{1}, Yi Han\textsuperscript{2}}

\IEEEauthorblockA{{\textsuperscript{1}School of Information Science and Technology, Dalian Maritime University, Dalian, China} \\
{\textsuperscript{2}R\&D Innovation Center, COSCO Shipping Technology Co., Ltd, Shanghai, China}\\
Email: xzy2408@dlmu.edu.cn, hellotem@hotmail.com}
}


\maketitle

\renewcommand\thefootnote{}
\footnote{† Corresponding Author}
\renewcommand\thefootnote{\arabic{footnote}}

\begin{abstract}
With the increasing demands for safety, efficiency, and sustainability in global shipping, Automatic Identification System (AIS) data plays an increasingly important role in maritime monitoring. AIS data contains spatial-temporal variation patterns of vessels that hold significant research value in the marine domain. However, due to its massive scale, the full potential of AIS data has long remained untapped. With its powerful sequence modeling capabilities—particularly its ability to capture long-range dependencies and complex temporal dynamics—the Transformer model has emerged as an effective tool for processing AIS data. Therefore, this paper reviews the research on Transformer-based AIS data-driven maritime monitoring, providing a comprehensive overview of the current applications of Transformer models in the marine field. The focus is on Transformer-based trajectory prediction methods, behavior detection, and prediction techniques. Additionally, this paper collects and organizes publicly available AIS datasets from the reviewed papers, performing data filtering, cleaning, and statistical analysis. The statistical results reveal the operational characteristics of different vessel types, providing data support for further research on maritime monitoring tasks. Finally, we offer valuable suggestions for future research, identifying two promising research directions. Datasets are available at https://github.com/eyesofworld/Maritime-Monitoring.
\end{abstract}


\section{Introduction}
The maritime industry is the backbone of global trade, handling over 80\% of international goods transportation \cite{brooks201850}. It connects continents, sustains economic growth, and ensures the movement of essential commodities, from crude oil and natural gas to consumer electronics and agricultural produce. With its critical role in supporting economies worldwide, the maritime industry faces increasing pressure to adapt to rapid global demands, efficiency, safety, and sustainability changes. In recent years, the maritime industry has undergone and is amid a significant transformation due to technological advancements and the increasing adoption of digital solutions \cite{raza2023digital}. Integrating digital technologies, such as AI and big data analysis, is revolutionizing operations ranging from cargo handling to fleet management. Besides, the industry is actively exploring revival strategies centered on technological innovation and sustainability \cite{terpsidi2019maritime}. Advancing the digitization of the industry through synergies between academic institutions and the shipping industry. By embracing these advancements, the maritime industry is better equipped to meet the challenges of the future and ensure the stability of the global trade enterprise.

Data-driven approaches are a key factor driving the digital transformation of the maritime industry \cite{leonard2019digitalization}. The Automatic Identification System (AIS) has become an essential component of maritime operations, providing real-time information on vessel location, speed, and other navigational details \cite{yang2024harnessing}. The widespread application of AIS has resulted in the generation of vast amounts of data, presenting both opportunities and challenges for maritime research \cite{fu2020ais}. AIS-driven solutions have significantly enhanced maritime safety and operational efficiency, from tracking vessels and predicting their trajectories to detecting predictive behaviors and optimizing energy efficiency.

However, the complexity of the marine environment and the dynamic nature of maritime activities demand that data analytics techniques effectively extract patterns from AIS data. AIS data includes static, dynamic, and navigational information, collected continuously over time, exhibiting time dependence and sequentiality, often represented as complex time series. While machine learning methods can capture temporal dependencies, they struggle to efficiently process large volumes of data. In contrast, deep learning techniques offer new avenues for handling the vast scale of AIS data \cite{li2023ais}. Among these, the Transformer model \cite{vaswani2017attention}, with its ability to capture long-range dependencies and complex temporal dynamics, has emerged as an effective solution for processing sequential data. The core algorithm of the Transformer, the attention mechanism, allows for the analysis of relationships between data points in a sequence, thereby uncovering intricate temporal patterns and dependencies, which are crucial for extracting key information from AIS data. As the maritime industry embraces digital transformation, the potential of Transformers in maritime monitoring has become increasingly prominent.

So far, a substantial body of review papers has discussed AIS data-driven maritime tasks, including vessel trajectory prediction, collision avoidance, anomaly detection, and energy efficiency management. For instance, Yang et al. \cite{yang2024harnessing} provided an overview of four primary application areas of machine learning in AIS data-driven maritime tasks. In trajectory prediction, Li et al. \cite{li2023ship} compared the applicability and performance of five classical machine learning methods and eight deep learning methods across different scenarios, offering guidance on selecting the appropriate prediction method for specific contexts. In the collision avoidance task, Rawson et al. \cite{rawson2023survey} emphasized the application of supervised machine learning, particularly in modeling vessel behavior, assessing collision risks, and providing automated collision avoidance decision support based on AIS data. In the anomaly detection task, Yan et al. \cite{yan2019study} discussed the application of statistical models, machine learning, and hybrid models. In contrast, Wolsing et al. \cite{wolsing2022anomaly} summarized five types of vessel anomalies and provided a systematic overview of their application scenarios. Regarding energy efficiency management, Yan et al. \cite{yan2021data} reviewed research progress in fuel consumption management, analyzing various fuel consumption prediction models and optimization strategies for vessels. On the other hand, Barreiro et al. \cite{barreiro2022review} conducted an in-depth exploration of the current state of vessel energy efficiency, optimization measures, and future research directions, providing valuable insights for researchers and practitioners in the field of vessel energy efficiency. These reviews provide deep insights into maritime tasks; however, studies on applying Transformer models in AIS data remain relatively scarce. Therefore, a comprehensive review of the application of Transformer models in AIS data-driven maritime monitoring is critical.

This review paper introduces the fundamental architecture and principles of AIS data and Transformer models and presents the AIS datasets we have collected and organized from the reviewed literature. Subsequently, we provide a comprehensive review of AIS data-driven maritime monitoring solutions based on Transformer models, focusing on three tasks: vessel trajectory prediction, vessel behavior detection, and vessel behavior prediction. We summarize and analyze the datasets and methods employed in existing research, highlighting the advantages of Transformer models in processing AIS data. We also discuss the solutions proposed in current studies for addressing various maritime tasks and scenarios. Overall, Transformer models provide a powerful tool for maritime monitoring that is capable of tackling most maritime tasks. This paper aims to offer a thorough review and analysis of AIS data-driven maritime monitoring tasks based on Transformer models, providing valuable insights for future research directions.

The remainder of the paper is structured as follows. Section 2 introduces the characteristics of AIS data, data cleaning methods, and the AIS dataset that we have collected and organized, as well as the application of the Transformer model in analyzing large-scale AIS data, highlighting its advantages in maritime monitoring tasks. Section 3 provides a comprehensive review of existing research on vessel trajectory prediction based on the Transformer model, including datasets, methods, and applicable scenarios, and discusses the applications of vessel trajectory prediction. Section 4 further reviews other maritime tasks, including vessel behavior detection and prediction, and briefly outlines relevant aspects of vessel energy management. Finally, Section 5 discusses and summarizes the work, offering suggestions for future research directions.

\begin{figure*}[htbp]
	\centerline{\includegraphics[width=\textwidth]{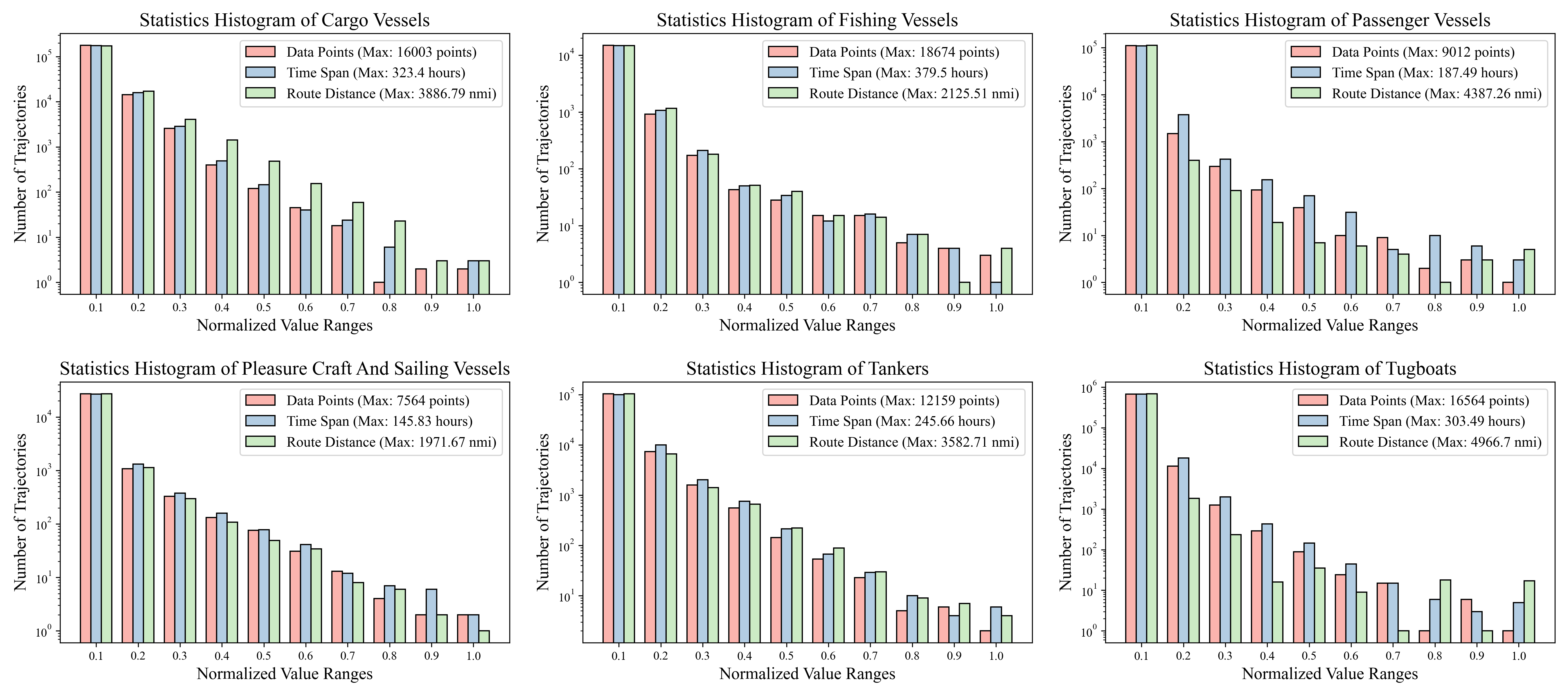}}
	\caption{Grouped statistical histograms of 6 types of vessels}
	\label{fig1}
\end{figure*}

\section{AIS Data \& Transformer}

\subsection{Automatic Identification system}

The Automatic Identification System (AIS) was initially developed to enhance radar functionality and vessel traffic services (VTS), providing identification and positioning information for vessels and shore stations. According to the International Convention for the Safety of Life at Sea (SOLAS), ships engaged in international voyages with a gross tonnage of 300 or more and non-international trading cargo ships with a gross tonnage of 500 or more are required to be equipped with AIS devices \cite{chircop2015international}. AIS broadcasts and receives positional information through Very High Frequency (VHF) radio waves, improving navigational safety and maritime traffic management. AIS data contains dynamic information (such as position and speed), static information (such as vessel type and name), and navigational information (such as destination port and estimated time of arrival) and has now become a crucial foundation for maritime task research.

However, raw AIS data often contains outliers and missing values, typically caused by VHF network fluctuations, communication channel congestion, and interference. Incorrect data can lead to erroneous conclusions in maritime data mining, misjudging vessel navigation patterns. To clean outliers, threshold-based filtering methods are commonly used, such as excluding implausible latitude and longitude values, speed over ground (SOG), and course over ground (COG). For handling missing values, interpolation methods (such as linear interpolation or spline interpolation) can be applied to estimate and fill in the missing data. Alternatively, machine learning or deep learning techniques can be employed for data imputation, where models are trained to analyze historical trajectories and predict the missing values.

For this study, we collected the publicly available datasets from the studies reviewed in this paper to the greatest possible extent and performed data cleaning to address any outliers, including those related to latitude and longitude, SOG, COG, and heading. Subsequently, we filtered the AIS data of vessels with more than 60 data points and a recording duration exceeding 1 hour and segmented the trajectories based on each vessel’s navigation records. This process resulted in a high-quality AIS dataset containing data from 19,016 vessels across six different ship types, comprising approximately 640 million AIS messages, which have been made publicly available.

Then, we summarized the number of data points, time span, and route distances for each vessel type, normalized the results and generated histograms for the grouped trajectories of these six ship types, as shown in Figure \ref{fig1}. The statistical results reveal a long-tail distribution, with most vessels concentrated below a normalized value of 0.1. Passenger ships, fishing boats, yachts, and sailboats primarily engaged in short-distance voyages, with their voyage distances and time spans concentrated in shorter ranges. In contrast, tankers, cargo ships, and tugboats tend to undertake longer voyages, characterized by extended durations and distances, highlighting their significant role in global transport or cross-regional operations. This pattern reveals the maritime behavior of different vessel types and provides valuable data support for further research on ship operation patterns and trajectory prediction.

\subsection{Transformer}
With the growth of global shipping activities and the widespread adoption of maritime traffic monitoring technologies, the volume of AIS data has surged, containing immense research potential \cite{wu2017mapping}. Extracting valuable information from this vast amount of AIS data has become one of the core challenges in the maritime field. In 2017, Vaswani et al. introduced the Transformer model, which has led a technological revolution in natural language processing due to its outstanding sequence modeling capabilities and computational efficiency. The Transformer model relies entirely on the self-attention mechanism and incorporates components such as multi-head attention, positional encoding, and feed-forward neural networks. Utilizing parallel processing and feature subspace analysis can efficiently extract complex dependencies in long time series, making it particularly well-suited for analyzing the dynamic attribute changes of vessels over extended periods.

Compared to traditional RNNs and LSTMs, the advantages of the Transformer are especially prominent. The core component of the Transformer, the self-attention mechanism, enables it to simultaneously attend to all positions in a sequence, overcoming the long-distance dependency issue common in RNNs and LSTMs. This is crucial for understanding vessel behavior patterns. Furthermore, the parallel computation ability of the Transformer significantly enhances processing efficiency, allowing it to excel in analyzing massive AIS datasets. The Transformer also offers flexibility in adjusting the model complexity, enabling it to capture complex temporal patterns and contextual information more accurately when processing long time series. As a result, the Transformer model is considered an ideal approach for analyzing vessel trajectories, behavior prediction, and other dynamic changes in AIS data.

\section{Vessel Trajectory Prediction}
Vessel trajectory prediction is widely recognized as one of the key technologies for achieving safe autonomous navigation, with the prediction methods and their performance serving as critical factors for the future of maritime safety and autonomous shipping \cite{li2023ais}. This section reviews Transformer-based AIS data-driven methods applied to trajectory prediction, categorizing them according to their prediction approaches. Finally, we introduce the related applications of trajectory prediction.

\begin{table*}[htbp]
	\caption{Summary of generative trajectory prediction methods}
	\begin{center}
		\begin{tabular}{@{} p{2.3cm} p{4cm} p{5cm} p{5.5cm} @{} }
			\hline
			\textbf{Reference} & \textbf{Data Volume} & \textbf{Method} & \textbf{Focused} \\
			\hline
			Billah et al. \cite{billah2022method} & 125K AIS data records & GRU, LSTNet, Transformer, LSTM, CNN & Short-term trajectory prediction \\
			Zhang et al. \cite{zhang2024dynamic} & 29,397 AIS data samples & iTransformer + GAN & Short-term trajectory and turning state prediction \\
			Li et al. \cite{li2023ais} & 16,185 vessel trajectories & Transformer & Short-term trajectory prediction \\
			Drapier et al. \cite{drapier2024enhancing} & 1,731,686 unique trajectories & Mixtral 8x7B & Long-term trajectory prediction \\
			Yuan et al. \cite{yuan4806937gatransformer} & 4985 ship trajectory sequences & Transformer + GAT & Complex navigable waters trajectory prediction \\
			Xue et al. \cite{xue2024g} & 12,808 vessel trajectories & Transformer + GRU & Ship trajectory and ship turning status prediction \\
			Donandt et al. \cite{donandt2024incorporating} & 264,272 ship trajectory sequences & Transformer + GMMs & Inland waterway navigation trajectory prediction \\
			Liu et al. \cite{liu2024spatio} & 3174 vessel trajectories & Transformer + GCN + RegLSTM & Multi-vessel trajectory prediction \\
			Xia et al. \cite{xia2023tatbformer} & 500 consecutive trajectories & Transformer + TCN & Long-term trajectory prediction \\
			Zhang et al. \cite{zhang2023trajectory} & 6,694 vessel trajectories & Transformer + GCN + TCN & Short-term and long-term trajectory prediction \\
			Jiang et al. \cite{jiang2023trfm} & No specific data size is mentioned & Transformer + LSTM & Short-term and long-term trajectory prediction \\
			Huang et al. \cite{huang2022tripleconvtransformer} & 336,573 vessel trajectories & Transformer + CNN & Short-term trajectory prediction \\
			Zhang et al. \cite{zhang2023nonlinear} & No specific data size is mentioned & Transformer & Short-term trajectory prediction \\
			Xiong et al. \cite{xiong2024informer} & 3,898 vessel trajectories & Informer & Multi-vessel long-term trajectory prediction  \\
			Jiang et al. \cite{jiang2023prediction} & 683 vessel trajectories & Transformer + GHC & Multi-vessel trajectory prediction \\
			Zou et al. \cite{zou2023dynamic} & Tens of thousands of ship tracks & Temporal Transformer & Multi-vessel trajectory prediction \\
			\hline
		\end{tabular}
	\end{center}
	\label{tab2}
\end{table*}

\begin{table*}[htbp]
	\caption{Summary of classification trajectory prediction methods}
	\begin{center}
		\begin{tabular}{@{} p{2.3cm} p{4cm} p{5.1cm} p{5.5cm} @{} }
			\hline
			\textbf{Reference} & \textbf{Data Volume} & \textbf{Method} & \textbf{Focused} \\
			\hline
			Nguyen et al. \cite{nguyen2024transformer} & 712 million AIS messages & TrAISFormer & Long-term trajectory prediction \\
			Donandt et al. \cite{donandt2022short} & 7.7K vessel trajectories & Context-Sensitive Classification Transformer & Inland vessels short-term trajectory prediction  \\
			Donandt et al. \cite{donandt2023improved} & About 117.5K vessel trajectories & Context-Sensitive Classification Transformer & Inland vessels trajectory prediction \\
			Sigillo et al. \cite{sigillo2023sailing} & 12,165,957 AIS messages & GPT models + Gumbel softmax & Long-term trajectory prediction \\
			Takahashi et al. \cite{takahashi2024ship} & No specific data size is mentioned & TrAISFormer & Fishing vessel trajectory prediction \\
			Xue et al. \cite{donandt2023spatial} & About 400k trajectory sequences & sosp-CT & Inland vessels short-term trajectory prediction \\
			\hline
		\end{tabular}
	\end{center}
	\label{tab3}
\end{table*}

\subsection{Generative Trajectory Prediction}
Generative trajectory prediction conceptualizes ship trajectory forecasting as a generative problem, emphasizing trajectories' continuity and dynamic variability. This approach generates a complete trajectory for a forthcoming period by learning the distribution of historical ship trajectories. A summary of generative trajectory prediction methods is presented in Table \ref{tab2}. We comprehensively review dataset sizes, methods, and focused tasks.

Each of the sixteen papers reviewed considered utilizing longitude, latitude, SOG, and COG from AIS data as inputs for their predictive models. Regarding prediction targets, three papers focused on multi-ship trajectory prediction (\cite{liu2024spatio,xiong2024informer,zou2023dynamic}), while one paper concentrated on trajectory prediction for large ocean-going bulk carriers \cite{huang2022tripleconvtransformer}. In terms of application scenarios, one study targeted trajectory prediction for inland navigation vessels \cite{donandt2024incorporating}, another dedicated to trajectory prediction in complex navigable waters \cite{yuan4806937gatransformer}, and a third focused on trajectory prediction in nearby port waters \cite{jiang2023prediction}. Such goal-specific studies can formulate tailored solutions based on the attributes of their respective targets, thereby achieving superior performance compared to general methods within their specific domains.

Generally, generative trajectory prediction methods exhibit broad applicability and perform commendably across various scenarios. They effectively address navigational uncertainties arising from unforeseen circumstances in trajectory predictions with relatively fixed routes and demonstrate exceptional performance in multi-ship trajectory prediction tasks.

\subsection{Classification Trajectory Prediction}
Classification trajectory prediction approaches ship trajectory forecasting as a classification problem. By analyzing historical trajectory data, these methods categorize future trajectory positions or states into discrete classes. These classes may represent specific regions, navigational channels, or ports, thereby enabling the prediction of a vessel's future trajectory. A summary of classification-based trajectory prediction methods is provided in Table \ref{tab3}.

In classification problems, the construction of classification labels is a critical component. Among the six papers reviewed, all constructed classification labels by discretizing continuous features and then partitioning these discretized features into intervals. For instance, Nguyen et al. \cite{nguyen2024transformer} divided latitude and longitude into intervals of 0.01\textdegree, SOG into 1-knot, and COG into 5\textdegree intervals. Additionally, three studies (\cite{donandt2023improved,donandt2022short,donandt2023spatial}) by the same author addressed the unique complexities of inland navigation by proposing customized solutions that consider multiple factors such as vessel behavior, spatial environment, and social context, thereby making significant contributions to inland vessel trajectory prediction. Another paper \cite{takahashi2024ship} focused on trajectory prediction for fishing vessels, whose navigation paths are typically more complex and variable than those of cargo ships and tankers and are significantly influenced by seasonal factors, thereby presenting greater challenges for trajectory prediction.

In summary, classification trajectory prediction methods are commonly employed for vessels navigating complex routes or those with high trajectory randomness. These seemingly straightforward methods demonstrate robust performance in complex scenarios by simplifying intricate trajectories into discrete categories and encapsulating the vessels' navigational experiences.

\subsection{Applications of Vessel Trajectory Prediction}
Vessel trajectory prediction is widely applied in maritime safety, traffic management, environmental protection, and smart shipping, among other fields. It plays a crucial role in collision prevention, emergency rescue, channel planning, and port management. Additionally, it enhances shipping efficiency, reduces emissions' impact on the marine ecosystem, and optimizes routes and autonomous ship navigation in intelligent shipping systems, thereby improving navigation safety.

Mandalis et al. \cite{mandalis2024transformer} proposed a distributed unified approach (dUA-VTFF) that uses Transformer models and the Apache Spark big data framework to predict vessel traffic flow for the next 30 minutes based on historical maritime data. Vessel traffic flow prediction impacts shipping efficiency and safety and is essential for environmental protection, port management, and smart shipping technologies. As shipping volume increases, the accuracy and timeliness of predictions become particularly critical, directly influencing global supply chains and the sustainable use of marine resources. Experimental results indicate that dUA-VTFF outperforms other methods in accuracy, efficiency, and scalability.

Gao et al. \cite{gao2024deep} applied vessel trajectory prediction to regional risk assessment and proposed a predictive framework capable of identifying high-collision-risk ship pairs and evaluating regional risks. The proposed method employs the DBSCAN algorithm to cluster ships with potential collision risks, defines hotspot areas, and then uses a Transformer network to predict the future positions of ships. The risk is assessed by combining DCPA (Dynamic Closest Point of Approach) and TCPA (Time to Closest Point of Approach). Based on AIS data from the Yangtze River in Jiangsu, the experiments demonstrate that the method can quantitatively assess the collision risk in hotspot areas and obtain overall risk through thresholding, making it applicable to risk assessments in other waterways. Regional risk assessment enhances traffic safety and efficiency in complex waterways, especially given the increasing traffic density in inland rivers and marine areas \cite{SILVEIRA2021107789}.

\section{Vessel Behavior Detection and Prediction}
This section focuses on the tasks of ship behavior detection, behavior prediction, and energy efficiency optimization. We review Transformer-based AIS data-driven methods applied to these tasks and analyze the reviewed studies.

\begin{table*}[htbp]
	\caption{Summary of vessel behavior detection methods}
	\begin{center}
		\begin{tabular}{@{} p{2.3cm} p{4.7cm} p{4.5cm} p{5.4cm} @{} }
			\hline
			\textbf{Reference} & \textbf{Data Source} & \textbf{Method} & \textbf{Behavior Description} \\
			\hline
			Zhang et al. \cite{zhang2023novel} & Water area of Yantai, China & Transformer Encoder & Abnormal ship position, speed and course \\
			Xie et al. \cite{xie2024anomaly} & Shanghai Wusong Port & Transformer + ProbSparse Attention & Abnormal ship position, speed and course   \\
			Min{\ss}en et al. \cite{minssen2024predicting} & German Bight, Elbe and Weser Rivers & Transformer Encoder & Abnormal ship trajectory in the waterway \\
			Li et al. \cite{li2023abnormal} & United States Coast Guard & Transformer Encoder & Abnormal trajectory \\
			Kong et al. \cite{kong2022aship} & European AIS dataset & Bayesian Transformer & Identify ship targets using track information \\
			Kong et al. \cite{kong2022bship} & Celtic Sea and Bay of Biscay (France) & Transformer & Identify ship targets using track information \\
			Gu et al. \cite{gu2024mfgtn} & Hainan's waters and the East China Sea & Transformer + Fast Attention & Identify single trawler trajectories \\
			Bernab{\'e} et al. \cite{bernabe2023detecting} & 4.05 billion AIS messages globally & Transformer Encoder & Intentional AIS Shutdown \\
			\hline
		\end{tabular}
	\end{center}
	\label{tab4}
\end{table*}

\begin{table*}[htbp]
	\caption{Summary of vessel behavior prediction methods}
	\begin{center}
		\begin{tabular}{@{} p{2.3cm} p{4.7cm} p{4.5cm} p{5.4cm} @{} }
			\hline
			\textbf{Reference} & \textbf{Data Source} & \textbf{Method} & \textbf{Behavior Description} \\
			\hline
			Sturgis et al. \cite{sturgis2024beyond} & United States Coast Guard & Transformer & Underway, moored, at anchor and drifting \\
			Zhang et al. \cite{zhang2023deep} & A typical Ro-Pax ship & Transformer & 6-DoF ship motion and ship turning radius   \\
			Zhang et al. \cite{zhang2024comparison} & A typical Ro-Pax ship & Transformer & 6-DOF motion of a ship along the track \\
			El et al. \cite{el2023deep} & Saint Lawrence Seaway & Transformer & Vessel speed prediction \\
			\hline
		\end{tabular}
	\end{center}
	\label{tab5}
\end{table*}

\subsection{Vessel Behavior Detection}
Vessel behavior detection is important for maritime safety and intelligent vessel management. Detecting anomalous behaviors helps ensure navigational safety, prevent accidents, and more effectively manage potential risks. It also plays a crucial role in enhancing vessel operational efficiency and optimizing ship safety monitoring systems. A summary of vessel behavior detection methods is presented in Table \ref{tab4}. We provide a brief overview of the dataset sources, methods, and description of the corresponding behaviors.

Of the eight papers reviewed, half detect abnormal vessel behavior by setting thresholds. Specifically, Zhang et al. \cite{zhang2023novel} used the Minimum Description Length (MDL) criterion to extract trajectory features, calculated trajectory similarity using an improved Dynamic Time Warping (DTW) algorithm, and employed DBSCAN clustering. They then applied a threshold-based detection method, training a Transformer model to predict normal trajectories and set alarm thresholds. Similarly, Xie et al. \cite{xie2024anomaly} used an improved DBSCAN for trajectory clustering, introduced the Adaptive Threshold DP algorithm with feature point selection (AF-DP), and employed Fast-DTW to compute trajectory similarity. They also enhanced the clustering threshold selection with an improved KANN and combined ProbSparse Attention with the Transformer model for short-term prediction. This paper introduced the concept of "ship behavior similarity" for the first time and set abnormal thresholds by analyzing velocity vectors. Min{\ss}en et al. \cite{minssen2024predicting} studied the detection of abnormal behavior of ships in waterways. The waterway was divided into grid cells, continuous trajectories were created using AIS data, and transition points (TP) were calculated. By linear interpolation and calculating the angle and distance of a ship entering a cell, they determined the threshold for abnormal behavior. The study also integrated weather and tidal data into their experiments. The results indicated that tidal data improved prediction accuracy, while weather data reduced model performance. The authors suggested this might be due to low-information data potentially introducing noise or bias.

Vessel target identification is also critical for ensuring maritime safety and obtaining battlefield situational awareness. Although AIS data provides information about vessel categories, it may be tampered with to conceal the true identity of the vessels. Two papers by Kong et al. \cite{kong2022aship, kong2022bship} studied identifying nine types of vessels. The former utilized a Bayesian-Transformer Neural Network (BTNN) to handle identification in high-noise environments, while the latter enhanced trajectory information by constructing a context knowledge base and integrating Transformer and LSTM models. Meanwhile, Gu et al. \cite{gu2024mfgtn} focused on improving the identification capabilities for single-trawl fishing vessels to reduce illegal fishing activities. By integrating AIS and radar data, they proposed the MFGTN model, which employs a dual-tower Transformer network, a Fast Attention module, and a Fusion Former visual module to optimize identification and process unstructured data.

To address the issue of discontinuous trajectories in vessel behavior detection, Bernab{\'e} et al. \cite{bernabe2023detecting} proposed a self-supervised deep learning method that uses a Transformer model to predict whether an AIS message should be received. The data processing splits the trajectory into historical messages and the most recent position, followed by normalization. During self-supervised training, the model determines whether an AIS message should be received within a specific time frame. Suppose the prediction indicates that no message is to be received but the actual situation requires one. In that case, the trajectory is flagged as abnormal, suggesting the possibility of intentional AIS shutdown, and an alert is issued to the operator.

\subsection{Vessel Behavior Prediction}
Vessel behavior prediction is crucial for enhancing the safety and efficiency of maritime navigation, especially as maritime traffic becomes increasingly complex and more advanced collision avoidance systems are required. Predicting ship movements helps understand potential risks, optimize route strategies, and improve maritime operations' overall management. A summary of the vessel behavior prediction methods is presented in Table \ref{tab5}, where we briefly outline the data source, methodology, and descriptions of the corresponding behaviors.

In the four papers reviewed, Sturgis et al. \cite{sturgis2024beyond} manually classify vessel behavior into four categories: underway, moored, anchored, and drifting, and predicts the behavior of the vessel at future timestamps based on these categories. The authors propose two distinct methods to eliminate geographical bias, aiming to consider ship behavior prediction on a global scale rather than within a limited geographical area. The first method enhances geospatial trajectory data with specific techniques, including adding angle information (e.g., direction) to reduce bias in the dataset. The second method designs geo-bias-free features by removing these biases from the underlying AIS characteristics. Additionally, this paper focuses directly on the classification of container ship behavior. Due to the influence of ports, container ship behavior is more susceptible to geographical bias.

In addition to predicting ship behavior, some studies directly predict ship motions in greater detail. Two papers by Zhang et al. \cite{zhang2023deep, zhang2024comparison} originate from the same research team. They have significantly contributed to developing intelligent ship control systems by predicting ship motions in 6-DoF under real-world conditions. In terms of data, they use AIS trajectories, hydrometeorological data, and seabed topography data to extract ship motion features and environmental conditions. In terms of methodology, they employ a Transformer-based deep learning approach combined with the 6-DoF FSI ship grounding dynamics model to predict ship motions under real conditions.

Among the reviewed papers, one article focuses on predicting dynamic ship attributes. El et al. \cite{el2023deep} studied the prediction of the speed required for a ship to reach the next reference point on its planned route in the VTS system. Although fixed speed values have been established between consecutive reference points in VTS, using the same values regardless of ship type, weather conditions, or route sequence can lead to inaccurate estimates. Therefore, this study aims to predict the speed required to reach the next reference point on the route, formulated as a supervised machine learning task.

\subsection{Vessel Energy Efficiency Management}
In the maritime domain, energy efficiency refers to the relationship between the energy consumed in ship transportation and the transport tasks \cite{ramsay2023maritime}, typically measured by fuel consumption or carbon dioxide emissions per unit of cargo transported \cite{wang2024improving}. Improving energy efficiency can reduce operational costs, decrease environmental pollution, and comply with international environmental regulations. With the development of the shipping industry and the increasing environmental requirements, traditional methods of measuring energy efficiency have gradually failed to meet the demands for real-time, accuracy, and automated optimization. Therefore, integrating AIS data with deep learning methods has become an effective approach to optimizing energy efficiency.

At the 80th session of the Marine Environment Protection Committee (MEPC), the International Maritime Organization (IMO) adopted the "2023 Ship Greenhouse Gas Emissions Reduction Strategy," which sets clear phased targets for reducing both the total and intensity of ship carbon emissions. To achieve this, the IMO proposed a series of short-term, medium-term, and long-term emission reduction measures, with a key short-term focus on reducing the carbon emission intensity of ships \cite{chuah2023implementation}. Li et al. \cite{li2023research} using the Port of Tianjin as a case study, developed a high-resolution ship carbon emission inventory and predicted future emissions based on AIS and Lloyd's data. The study employed a top-down dynamic approach to calculate the carbon emissions of vessels. It analyzed real-time emissions from different port facilities using geographic information of the Port of Tianjin. Based on this, a Temporal Fusion Transformer (TFT) model was used to forecast vessels' multi-period carbon emission characteristics. The findings provide a scientific basis for environmental management and emission reduction policies at the Port of Tianjin and other global ports.

\section{Conclusion}
This study comprehensively reviews the latest advancements in AIS data-driven maritime monitoring research based on Transformer models, focusing on vessel trajectory prediction, behavior detection, and behavior prediction tasks. The goal is to highlight the potential of combining Transformer models with AIS data in the maritime domain. For the reviewed papers, we discuss their respective datasets, methodologies, and areas of emphasis. Vessel trajectory prediction methods are classified according to their generation methods, while other maritime dynamic monitoring tasks are unified under behavior detection and prediction. We also collected and organized publicly available datasets from the papers, creating a high-quality AIS dataset after cleaning and filtering the data. A separate statistical analysis was conducted for different vessel types, and the results reveal the characteristics of different vessel types in maritime behavior, providing valuable data support for further research on vessel operation patterns. Meanwhile, Transformer models offer a powerful tool for analyzing AIS data.

Future research in maritime monitoring tasks can develop in two main directions. The first is an improvement in data quality. Raw AIS data is frequently affected by noise during collection and transmission, leading to outliers or missing values. Consequently, the precise reconstruction of vessel trajectories has become a crucial aspect of maritime research. This reconstruction can be approached through both mathematical and learning methods. Mathematical methods, such as interpolation and fitting, primarily use mathematical functions to focus on the geometric features of vessel trajectories, but they lack the ability to analyze vessel behavioral patterns. On the other hand, learning methods like RNNs and LSTMs require large amounts of high-quality historical data for training. Exploring new AIS data preprocessing techniques for reconstructing vessel trajectories is of significant importance for the success of maritime monitoring tasks. In addition, integrating multi-source data is another promising direction for development. For example, meteorological data can provide environmental information, and radar data is often more reliable than AIS in areas such as channels or ports. However, multi-source data is often heterogeneous, which requires more sophisticated preprocessing methods and models with enhanced feature extraction capabilities.

The second direction involves improvements in methods. Since the development of the Transformer model, researchers have proposed numerous modifications. We no longer need to be confined to the standard Transformer model but should explore the feasibility of its variants in maritime task research. Additionally, the use of hybrid models represents another promising avenue. For instance, employing graph neural networks can capture interactions between vessels, and combining this with Transformer models can enhance multi-vessel trajectory prediction. In summary, the potential of Transformer-based AIS data-driven maritime monitoring technologies goes far beyond current applications. We believe this technology will be increasingly important in smart shipping and maritime safety management, driving the maritime industry toward more efficient, safer, and sustainable operations.

\end{document}